\documentclass{article}
\usepackage{graphicx}
\usepackage{times}
\usepackage{mathptmx}

\usepackage{url}
\usepackage{algorithm}
\usepackage{algorithmic}

\title{Text Detection on Roughly Placed Books by Leveraging a Learning-based Model Trained with Another Domain Data\thanks{This paper is based on a previous paper published in the proceedings of the 4th IEEE Cyber Science and Technology Congress (CyberSciTech), 2019~\cite{cyber}}}
\author{Riku Anegawa$^1$ and Masayoshi Aritsugi$^2$\\
  {\small\em $^1$ Computer Science and Electrical Engineering} \\
  {\small\em Graduate School of Science and Technology}, 
  {\small\em Kumamoto University}\\
  {\small\em $^2$ Big Data Science and Technology}\\
  {\small\em Faculty of Advanced Science and Technology},
  {\small\em Kumamoto University}\\
  {\small\em 2-39-1 Kurokami, Chuo-ku, Kumamoto 860-8555, Japan}\\
  {\small\em E-mail: \{ anegawa@dbms., aritsugi@ \}cs.kumamoto-u.ac.jp}  
}

\begin{document}

\maketitle

\begin{abstract}
Text detection enables us to extract rich information from images.
In this paper, we focus on how to generate bounding boxes that
are appropriate to grasp text areas on books
to help implement automatic text detection.
We attempt not to improve a learning-based model by training it with an enough amount of data
in the target domain but to leverage it, which has been already trained with
another domain data.
We develop algorithms that construct the bounding boxes
by improving and leveraging the results of a learning-based method.
Our algorithms can utilize different learning-based approaches
to detect scene texts.
Experimental evaluations demonstrate that our algorithms work well
in various situations where books are roughly placed.
\end{abstract}

\section{INTRODUCTION}
It is required to automatically manage a lot of products, such as books and packing boxes.
Although some additional equipment including bar codes and RFID tags is often used
for the management~\cite{rfid}, it may require us to cost too much.
Some texts are supposed to be printed on any products to be managed,
and this study intends to detect the texts from an image of products automatically and use for management.
In this paper, we focus on books, which are roughly located in bookshelves,
and attempt to detect texts on them.

There have been many studies of identifying books and text detection on them~\cite{wci,jcdl,spine,smart}.
For example, \cite{smart} exploited line extractions based on the Hough transform.
However, such work may fail to find text areas exactly when a book has a title of many words. 
Recently, text detection from natural scenes has intensely been developed as learning-based methods~\cite{east,r2cnn,feature,reading,s1,s2,s3}.
We consider leveraging learning-based methods for detecting texts in natural scene images in the text detection on books.

Note that we do not attempt to improve the learning-based methods by training them with a good set of data
but adopt them with additional refinement processes.
This intention comes from the fact that it is often difficult
to collect enough amount of data in the target domain
and make them to be used for training.
Recently, transfer learning has intensely been studied~\cite{survey1,survey2}.
Transfer learning is the technology that re-learns only a part of the parameters of a model trained in another domain.
In this paper, we exploit trained models instead of re-learning.
We develop algorithms that construct bounding boxes for text areas by improving
and leveraging the results of a learning-based method.

\begin{figure}
\centerline{\includegraphics[width=123mm]{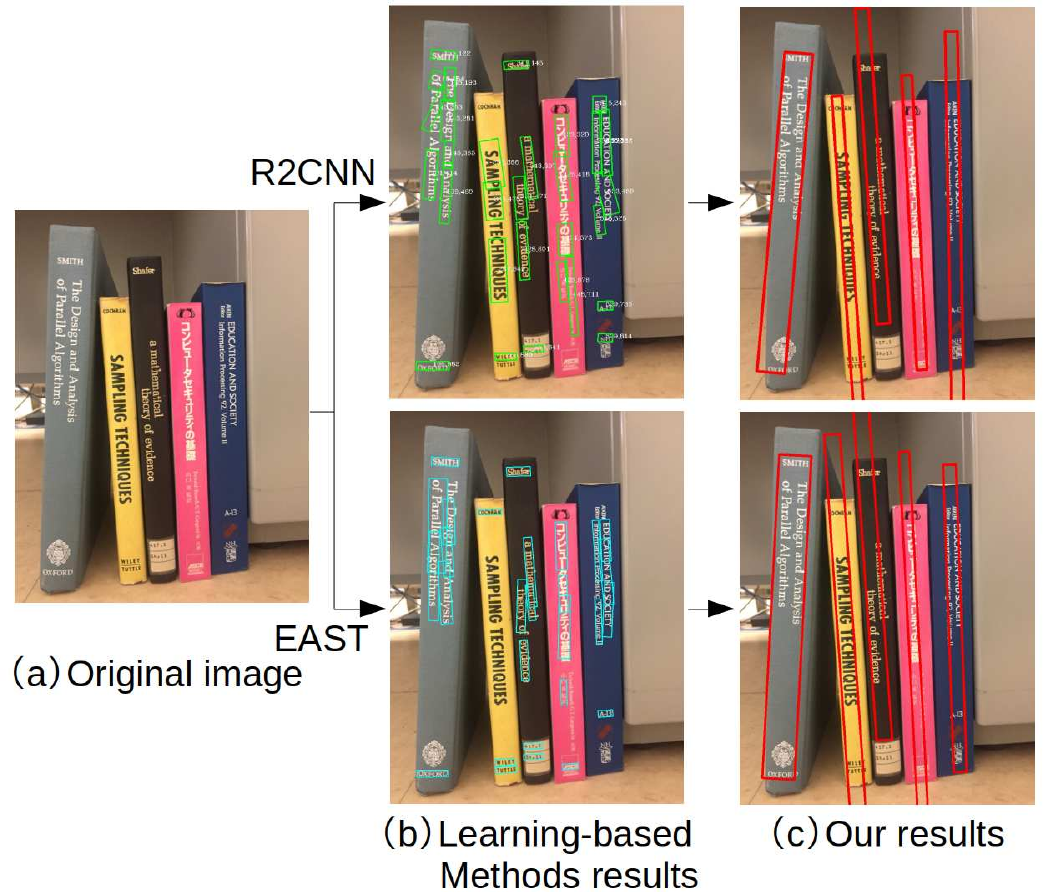}}
\caption{Overview of this study.}
\label{img-intro}
\end{figure}

This paper is based on our previous work~\cite{cyber}, in which
a convolutional neural network (CNN)-based method called R$^2$CNN~\cite{r2cnn}
was used to develop concrete discussions.
In this paper, we additionally use another learning-based method called EAST~\cite{east},
which is based on a fully convolutional network model and achieves a fast and
accurate scene text detection pipeline.
Figure~\ref{img-intro} shows an overview of our study where an image is processed
first by a learning-based method, i.e., R$^2$CNN or EAST,
and the result is then processed by our proposal to get the final result
of text detection.

The main contributions of this paper are as follows:
\begin{itemize}
\item First, we develop the discussion by adopting not only R$^2$CNN~\cite{r2cnn} but also EAST~\cite{east},
which is different from that used in our previous paper~\cite{cyber}, to show that our proposal
can utilize different learning-based text detection approaches without any modification
to them and improve their performances.  Our proposal can be of benefit to general learning-based approaches.
\item Second, we treat colors of spine and text strings on a book as binary data to give better performance than our previous algorithms
where they were treated as color histograms.
\item Third, we evaluate our proposed method with images where books were placed not only roughly but also horizontally on a bookshelf
to demonstrate how well our algorithms work in general situations.
\end{itemize}

The remainder of this paper is organized as follows.
Section~\ref{rel} mentions related work.
Section~\ref{prop} describes the details of our algorithms,
and the results of experimental evaluation are presented in Section~\ref{exp}.
Finally, our conclusions and directions for future research are presented in Section~\ref{concl}.

\section{RELATED WORK}\label{rel}
There have been studies of book management.
Shu et al.~\cite{rfid} proposed an accurate location algorithm based on an RFID intelligent bookshelf.
Although the proposed algorithm improved the positioning accuracy,
it naturally supposed a number of RFIDs were available, which can be costly.
Chen et al.~\cite{smart} developed a mobile book recognition system
which took photos of a bookshelf and exploited them for the management.
The system achieved automatic and robust book recognition.
However, the system assumed that almost all books were neatly arranged on the bookshelf
because their line extractions were based on the Hough transform.
In this paper, we consider cases where books are not always arranged very carefully
as shown in Fig.~\ref{overview}(a).

In recent years, scene text detection has been studied~\cite{east,r2cnn,feature,reading,s1,s2,s3},
which are learning-based approaches.
Jiang et al.~\cite{r2cnn} proposed an arbitrary-oriented scene text detection method
called Rotational Region CNN (R$^2$CNN), which was based on Faster R-CNN~\cite{fast}.
We described concrete discussions with using R$^2$CNN in our previous paper~\cite{cyber}.
Zhou et al.~\cite{east} proposed a fast and accurate scene text detection pipeline utilizing a fully convolutional network model.
In this paper, we develop algorithms which can generate good detection results of texts on books by leveraging a learning-based approach.
We use the studies of \cite{r2cnn} and \cite{east} as concrete methods.
Both of the learning-based models we used in this study were trained with ICDAR 2015 data~\cite{icdar}, which were different from our target application domain.

\section{REFINEMENTS OF THE RESULTS OF LEARNING-BASED APPROACHES}\label{prop}
\subsection{Results of Text Detection on Books by Learning-based Models Trained with Another Domain Data}
R$^2$CNN~\cite{r2cnn} generates region proposal for each CNN feature map by using the Region Proposal Network (RPN).
For each box generated by RPN, three ROI Poolings with different pooled sizes, namely, 7$\times$7, 3$\times$11, and 11$\times$3, are performed.
While Faster R-CNN~\cite{fast} uses ROI Poolings with pooled size of 7$\times$7 only,
R$^2$CNN does the poolings with the three sizes to handle horizontal or vertical texts.
The pooled features are then concatenated, and text/non-text scores,
axis-aligned boxes and inclined minimum area boxes are predicted.
The inclined boxes are processed by inclined non-maximum suppression
and the detection results are finally generated.

Zhou et al.~\cite{east} developed an efficient and accurate scene text detector (EAST).
They proposed a two-stage pipeline utilizing a fully convolutional network model
for predicting rotated rectables or quadrangles for text regions by incorporating
proper loss functions.

R$^2$CNN and EAST we used in this work were trained with ICDAR 2015 data~\cite{icdar}, which were
in a different domain from our target, i.e., images of roughly placed books.
An application way of such a learning-based approach is to train it with a considerable amount of data in the target domain.
Note, however, that we cannot suppose that enough amount of data in the target application domain
were always available.

Figure~\ref{overview}(b) shows text detection results of applying the two approaches to Figure~\ref{overview}(a),
in which books are located in a bookshelf in a natural way.
While both could almost successfully detect arbitrary-oriented texts on books,
they tend to detect each word individually.
As shown in the figure, there are roughly two problems in the results.
One is that bounding boxes may not cover texts correctly.
The other is that there are more than one bounding box detected on a book in most cases.
They would be mainly because the learning-based models were trained with different domain data.

\subsection{Refinement Algorithms}
\subsubsection{Overview}
As shown in Figure~\ref{img-intro}(b), the results of naive application of learning-based approaches
are not satisfied for the text det{{ection on books.
We therefore propose a refinement process in this paper.
Algorithm~\ref{overview} shows an overview of the text detection process.

\begin{algorithm}
  \caption{$Text Detection Process$($image$)}
  \label{overview}
  \begin{algorithmic}[1]
  \REQUIRE $image$
  \ENSURE $box\_groups$

  \STATE $boxes \leftarrow SceneTextDetection(image)$ \label{point1}
  \STATE Perform a Gaussian filter to $image$. \label{gaussian}
  \STATE Sort $boxes$ with their center coordinates from the top to the bottom of  $image$. \label{sort}
  \STATE Calculate $spine\_colors$ and $text\_colors$ of $boxes$. \\ \COMMENT{They are used in Algorithms~\ref{adjusting} and \ref{grouping}.} \label{cpoint}
  \STATE $boxes \leftarrow adjusting(image\&boxes, ~'location', ~adjust\_range)$ \label{al}
  \STATE $boxes \leftarrow adjusting(image\&boxes, ~'angle', ~adjust\_range)$ \label{aa}
  \STATE $flag \leftarrow True$
  \STATE $\#boxes \leftarrow 0$
  \WHILE{$flag$}
    \STATE $box\_groups \leftarrow grouping(image\&boxes, ~wide\_range\_rate)$ \label{gpoint}
    \IF{$\#boxes$ == $len(box\_groups)$}
      \STATE $flag \leftarrow False$
    \ELSE
      \STATE $\#boxes \leftarrow len(box\_groups)$
    \ENDIF
  \ENDWHILE \label{end}
  \STATE Perform the non-maximum suppression to $box\_groups$ \label{nms}
  \STATE Remove too small bounding boxes from $box\_groups$ \label{last}
  \RETURN $box\_groups$
  \end{algorithmic}
\end{algorithm}

A learning-based scene text detection approach is utilized at line~\ref{point1} in the algorithm.
Note that there is no need to modify, customize, nor re-learn it in our proposal.
We perform a Gaussian filter to the image (line~\ref{gaussian}).
At line~\ref{sort} we sort all bounding boxes obtained by the learning-based approach
in spatial order.
Then, we perform our proposed refinement process(es) between lines~\ref{cpoint} and \ref{end}.
Finally, we perform the non-maximum suppression to get the results (line~\ref{nms}) and
remove too small bounding boxes from the results (line~\ref{last}).

As in Algorithm~\ref{overview}, we propose two processes in this paper: adjusting (lines~\ref{al} and \ref{aa}) and grouping (line~\ref{gpoint}).
They respectively correspond to the problems appearing in the results of learning-based approaches, i.e., bounding boxes may not cover texts correctly
and there are more than one bounding box detected on a book in most cases.
We describe the details in the following.

\subsubsection{Book colors}
We assume that there are two colors on a book: spine and text string colors.
We apply k-means with $k=2$ on a bounding box.
Also, we assume that the four corners of a bounding box are almost occupied by the spine color, and thus the majority color of the four corners is set as the provisional spine color.
Then, out of the two colors the one similar to the provisional spine color is set as the spine color and the other is as the text string color.
Figure~\ref{text-spine} shows examples of extracted colors.

\begin{figure}
\centerline{\includegraphics[width=99mm]{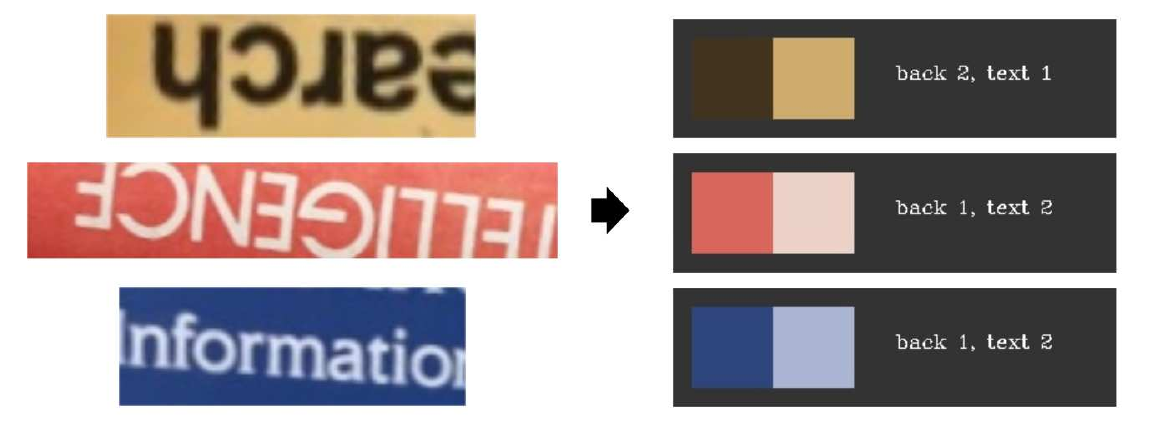}}
\caption{Spine and text colors.}
\label{text-spine}
\end{figure}

While we treated colors as color histograms in our previous paper~\cite{cyber},
we decided to treat them as binary data.
The binary treatment allows
us to have robust handling on colors by ignoring subtle varieties of colors appearing on images due to some situations,
e.g., light and book arrangement conditions.
This effect can be observed in the results of the experiments in Section~\ref{exp}.

\subsubsection{Adjusting}
As shown in Fig.~\ref{img-intro}, there are bounding boxes that do not cover texts correctly.
However, a bounding box generated by a learning-based approach does not completely
fail to cover texts.
In other words, the problem could be solved by slightly moving bounding boxes.
We consider approaching this problem by adjusting their locations and/or angles.
The algorithm is described in Algorithm~\ref{adjusting}.

\begin{algorithm}
  \caption{$adjusting$($image\&bounding\_boxes$, flag, $adjust\_range$)}
  \label{adjusting}
  \begin{algorithmic}[1]
  \REQUIRE $image\&bounding\_boxes$ - image with sorted bounding boxes in spatial order
  \ENSURE $adjusted\_boxes$

  \STATE $adjusted\_boxes \leftarrow  \emptyset$
  \FOR{$i=0$ to $len(bounding\_boxes)$}
    \STATE $box \leftarrow bounding\_boxes[i]$
    \STATE $max\_ret\_sum = -\infty$

	\IF{flag == 'location'}
	\STATE $adjust\_range$ stands for the range in pixel
	\ELSE
	\STATE $adjust\_range$ stands for the range in degree of angle
	\ENDIF

    \FOR{{\bf each} $i$ in $adjust\_range$}
      \STATE $shifted\_box \leftarrow shift(box, i)$
      \STATE $ave\_color \leftarrow$ calculate the average color of $shifted\_box$
      \STATE $norm\_spine \leftarrow norm(ave\_color - box.spine\_color)$
      \STATE $norm\_text \leftarrow norm(ave\_color - box.text\_color)$
      \STATE $norm\_sum \leftarrow norm\_spine - norm\_text$
      \IF{$norm\_sum > max\_ret\_sum$}
        \STATE $max\_ret\_sum \leftarrow norm\_sum$
        \STATE $best\_box \leftarrow shifted\_box$
      \ENDIF
    \ENDFOR
    \STATE $adjusted\_boxes$.append($best\_box$)
  \ENDFOR
  \RETURN $adjusted\_boxes$
  \end{algorithmic}
\end{algorithm}

In the algorithm, $shift()$ changes the location or the angle of a box
slightly in order to look for its more appropriate location or angle,
and $norm()$ returns the Euclidean distance.
We also shrink the bounding box in $shift()$ as 0.85 and 0.6 sizes in $x$ and $y$ axes of its original size
to take account of the effects of rotation and of the fact that the $y$ size is much larger than the $x$ size of a book.
The value of $adjust\_range$ is treated in pixel or angle value according to the given flag as in Algorithm~\ref{overview}.
An example of adjusting is shown in Fig.~\ref{adj-exa}.

\begin{figure}
\centerline{\includegraphics[width=108mm]{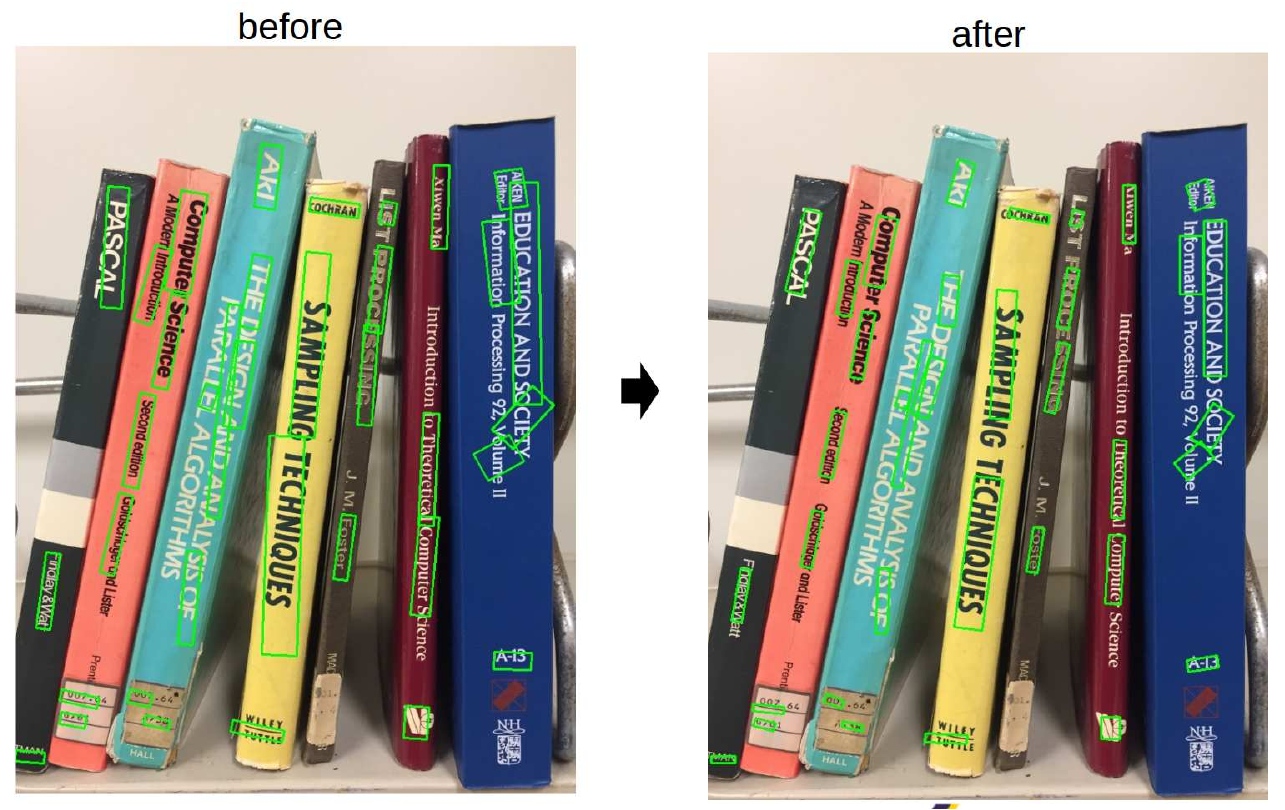}}
\caption{Adjusting results.}
\label{adj-exa}
\end{figure}

\subsubsection{Grouping}
We assume that detected bounding boxes of a book should be located closely and have angles similar to that of the book.
At line~\ref{sort} in Algorithm~\ref{overview} we get the top bounding box of each book.
Then, we calculate the range of each book with the top bounding box.
Figure~\ref{book-range} shows an overview of the calculation where the range has its center $(range.x, range.y)$.
We assume that the angles and the widths of the top bounding box and its book are the same, that is, $range.angle = box.angle$ and $range.width = box.width$.
We calculate the range as follows:
\begin{eqnarray}
box\_w & = & box.height \times cos(box.angle) + box.width \times sin(box.angle) \\
h & = & \frac{box\_w}{2} \times tan(box.angle) \\
line\_h & = & h + (img.height - box.y) \\
line\_w & = & \frac{line\_h}{tan(box.angle)} \\
range.height & = & \frac{line\_h}{sin(box.angle)} \\
range.x & = & box.x + \frac{box\_w}{2} - \frac{line\_w}{2} \\
range.y & = & box.y - \frac{box\_w}{2} \times tan(box.angle) + \frac{line\_h}{2}
\end{eqnarray}

\begin{figure}
\centerline{\includegraphics[width=88mm]{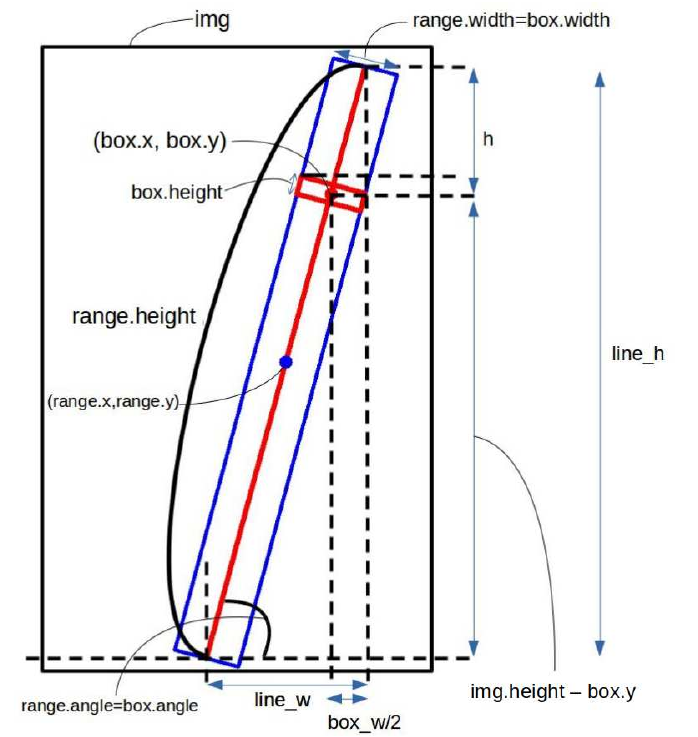}}
\caption{Book range.}
\label{book-range}
\end{figure}

We assume that two bounding boxes on a book should have similar colors of spine and text strings.
We calculate the similarities of a candidate pair of bounding boxes in terms of colors and locations
and combine them if the similarities are less than thresholds.
The algorithm of our grouping process is described in Algorithm~\ref{grouping} where
$wide\_range\_rate$ is used for relaxing the range of calculated book.

\begin{algorithm}
  \caption{$grouping$($image\&bounding\_boxes$, $wide\_range\_rate$)}
  \label{grouping}
  \begin{algorithmic}[1]
  \REQUIRE $image\&bounding\_boxes$ - image with sorted bounding boxes in spatial order, $wide\_range\_rate$ - which is multiplied to $range$
  \ENSURE $box\_groups$

  \STATE $box\_groups \leftarrow \emptyset$
  \WHILE{$bounding\_boxes \neq \emptyset$}
    \STATE $box\_group \leftarrow \emptyset$
    \STATE $box1 \leftarrow bounding\_boxes[0]$
    \STATE $box\_group.$append$(box1)$
    \STATE $range \leftarrow$ book range calculated as in Fig.~\ref{book-range}
    \STATE $range.width \leftarrow range.width \times wide\_range\_rate$
    \FOR{$i=0$ to $len(bounding\_boxes)$}
      \STATE $box2 \leftarrow bounding\_boxes[i]$
      \STATE $spatial\_flag \leftarrow False$
      \IF{More than 50\% of box2 is included in $range$}
        \STATE $spatial\_flag \leftarrow True$
      \ENDIF

      \STATE $norm\_spine \leftarrow norm(box1.spine\_color - box2.spine\_color)$
      \STATE $norm\_text \leftarrow norm(box1.text\_color - box2.text\_color)$
      \IF{$norm\_text < threshold\_text ~{\bf and} ~norm\_spine < threshold\_spine$}
        \STATE $color\_flag \leftarrow True$
      \ELSE
        \STATE $color\_flag \leftarrow False$
      \ENDIF

      \IF{$spatial\_flag$ {\bf and} $color\_flag$}
        \STATE $box\_group$.append($box2$)
      \ENDIF
    \ENDFOR
    \STATE $bounding\_boxes \leftarrow bounding\_boxes\setminus box\_group$
    \STATE $box\_groups \leftarrow box\_groups \cup box\_group$
  \ENDWHILE
  \RETURN $box\_groups$
  \end{algorithmic}
\end{algorithm}

\section{EXPERIMENTS}\label{exp}
We compare our proposal with R$^2$CNN~\cite{r2cnn}, EAST~\cite{east} and our histogram-based refinements~\cite{cyber}
with images of two types in terms of book arrangement: books were placed roughly and horizontally on a bookshelf.
In R$^2$CNN the anchor scales were set $(4, 8, 16, 32)$.
We used resnet-50 in EAST while PVANET was used in \cite{east}.
They were trained with using ICDAR 2015 data~\cite{icdar}.
We used a $5 \times 5$ Gaussian filter.
We set $adjust\_range=\pm{5}$ and $adjust\_range=\pm{10}$ in pixel and degree of angle, respectively, 
$threshold\_text=100$, $threshold\_spine=60$, and $wide\_range\_rate=2$,
In all the methods, bounding boxes with edges, each of which was less than or equal to 100 pixels
were removed at line~\ref{last} in Algorithm~\ref{overview}.

We compare the methods using four metrics, namely,
box accuracy (BA), error distance between centers (EDBC), intersection over union (IoU), and
area difference mean (ADM).
BA is defined as follows:
\begin{equation}
BA = \frac{\# books \ each \ of \ which \ has \ one \ bounding \ box}{\# books}.
\end{equation}
We explain the other metrics with Fig.~\ref{iou} where g is a ground truth, b is
the calculated bounding box, and area(box) and center(box) return box's area and center,
respectively.  The three metrics are defined as follows:
\begin{eqnarray}
EDBC(b, g) &=& norm(center(b) - center(g)) \\
IoU(b, g) &=& \frac{area(b \cap g)}{area(b \cup g)} \\
ADM(b, g) &=& abs(area(b) - area(g))
\end{eqnarray}
where $norm()$ and $abs()$ return the Euclidean distance and absolute value, respectively.
Note that these three metrics are calculated for books each of which was assigned one bounding box, that is, those counted in the numerator of BA.

\begin{figure}
\centerline{\includegraphics[width=50mm]{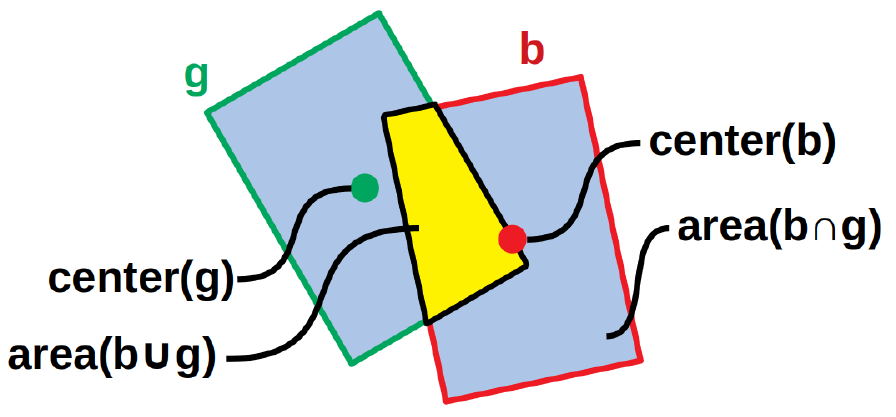}}
\caption{Relation between two bounding boxes.}
\label{iou}
\end{figure}

In the experiments, the ground truths were generated manually.

\begin{figure}
\centerline{\includegraphics[width=123mm]{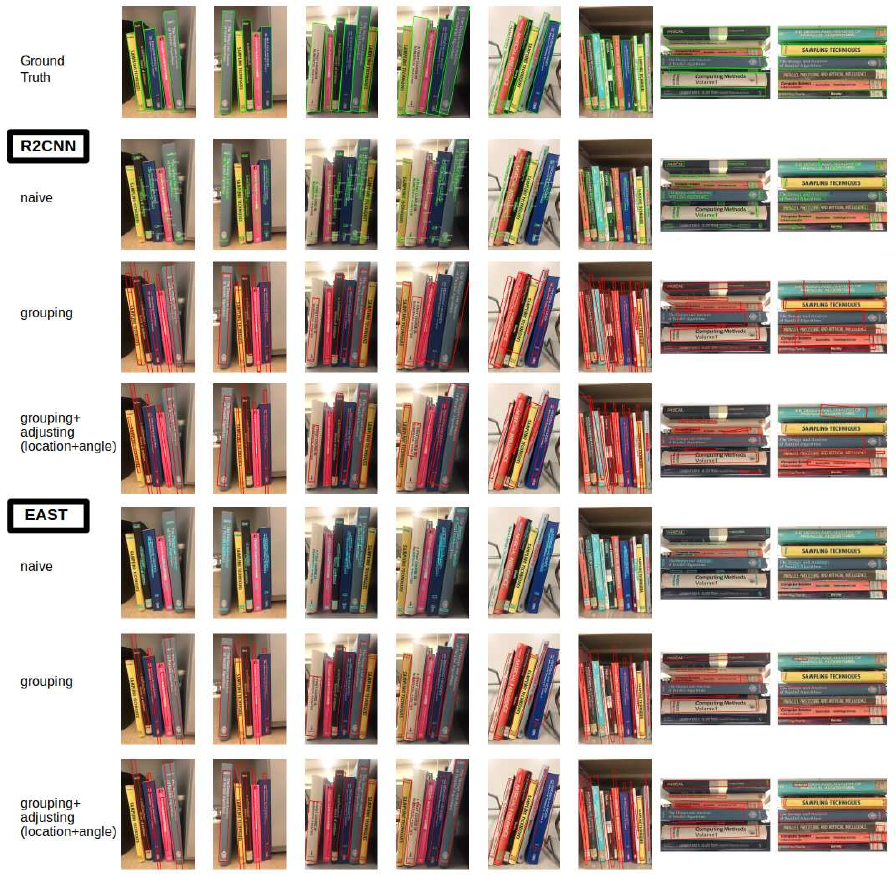}}
\caption{Results.}
\label{img-table}
\end{figure}

Figure~\ref{img-table} shows some obtained results of naively applications of the two methods,
grouping only and grouping+adjusting(location+angle) of our proposal.
First 6 and the rest 2 columns show the results with images of books which were placed roughly and horizontally, respectively.

\subsection{Roughly Placed Books}\label{rough}
We took 20 images of books roughly placed in bookshelves with various light conditions.
11 images were $1108 \times 1478$ and the rest 9 images were $890 \times 1188$ pixels.
There were 27 books in total in the images.

\begin{table}
  \begin{center}
  \caption{Roughly placed books cases.}
      \label{tab1}
    \scalebox{0.84}{

    \begin{tabular}{|c|c|c|c|c|c|c|} \hline
      \multicolumn{3}{|c|}{}&BA&EDBC&IoU&ADM\\\hline
      R2CNN&&naive&0.4609&115.33&0.2038&44698.2 \\ \cline{2-7}
      &histogram&grouping&0.8047&65.38&0.4381&31969.0 \\ \cline{3-7}
      &&grouping+adjusting(location)&0.7109&61.0338&0.3476&26899.7 \\ \cline{3-7}
      &&grouping+adjusting(location+angle)&0.5703&78.34&0.2838&40638.6 \\ \cline{2-7}
      &color clustering&grouping&0.7969&53.87&0.2842&41461.7 \\ \cline{3-7}
      &(our proposal)&grouping+adjusting(location)&0.7891&\textbf{51.19}&0.4355&\textbf{26490.6} \\ \cline{3-7}
      &&grouping+adjusting(location+angle)&\textbf{0.8125}&63.01&\textbf{0.4402}&34432.8 \\ \hline \hline
      EAST&&naive&0.3750&196.85&0.2994&75492.0 \\ \cline{2-7}
      &histogram&grouping&0.6953&74.89&0.4530&53990.9 \\ \cline{3-7}
      &&grouping+adjusting(location)&0.6953&73.68&0.4553&52627.4 \\ \cline{3-7}
      &&grouping+adjusting(location+angle)&0.7266&78.09&0.44270&56930.0 \\ \cline{2-7}
      &color clustering&grouping&0.7031&\textbf{70.79}&\textbf{0.4840}&\textbf{50354.0} \\ \cline{3-7}
      &(our proposal)&grouping+adjusting(location)&0.7188&78.51&0.4724&51890.2 \\ \cline{3-7}
      &&grouping+adjusting(location+angle)&\textbf{0.7344}&73.72&0.4708&51845.0 \\ \hline
    \end{tabular}
    }
  \end{center}
\end{table}

Table~\ref{tab1} shows the results of the comparison with images of roughly placed books.
Our proposal outperformed naive applications of R$^2$CNN and EAST.
Also, our proposal could give better performance than our previous paper.
In particular, our grouping+adjusting(location+angle) gave the best BAs,
meaning that our proposal could give more correct bounding boxes than the others.
While our histogram-based refinements sometimes failed to group bounding boxes belonging to a book
into one bounding box, our proposal of this paper was able to recognize different colors better than our histogram-based refinements.
Comparing adjusting results of our histogram-based refinements and our proposal of this paper,
we can say that the color handling of this paper made our adjusting process work well.

The results also indicate that the application of R$^2$CNN would be better than that of EAST.
This is mainly because sometimes EAST could not detect any bounding box on a book.
We should note that, while our proposal could improve the performance of a learning-based approach,
the performance of the final results would depend on the performance of the base learning-based approach.

\subsection{Horizontally Placed Books}
We took 15 images of books which were placed horizontally in bookshelves with various light conditions.
The images were $1108 \times 1478$ pixels and there were 27 books in total in the images.
The books were placed disorderly, and thus some were overhung by others.
We rotated the images 90 degrees and processed them in the same way as described in Section~\ref{rough}.

\begin{table}
  \begin{center}
  \caption{Horizontally placed books cases.}
      \label{tab2}
    \scalebox{0.84}{
    \begin{tabular}{|c|c|c|c|c|c|c|} \hline
      \multicolumn{3}{|c|}{}&BA&EDBC&IoU&ADM\\\hline
      R2CNN&&naive&0.2381&153.53&0.30872&45832.4 \\ \cline{2-7}
      &histogram&grouping&0.8476&71.33&0.48704&33451.2 \\ \cline{3-7}
      &&grouping+adjusting(location)&0.75238&74.99&0.49449&35851.2 \\ \cline{3-7}
      &&grouping+adjusting(location+angle)&0.67619&109.72&0.26503&57451.0 \\ \cline{2-7}
      &color clustering&grouping&0.85714&\textbf{61.82}&0.53103&\textbf{31633.9} \\ \cline{3-7}
      &(our proposal)&grouping+adjusting(location)&\textbf{0.87619}&62.71&\textbf{0.53178}&32453.9 \\ \cline{3-7}
      &&grouping+adjusting(location+angle)&0.82857&73.03&0.29235&39160.1 \\ \hline \hline
      EAST&&naive&0.2422&346.98&0.2208&163013.8 \\ \cline{2-7}
      &histogram&grouping&0.7238&165.76&0.4183&101556.4 \\ \cline{3-7}
      &&grouping+adjusting(location)&0.7238&165.89&0.4164&101563.5 \\ \cline{3-7}
      &&grouping+adjusting(location+angle)&0.6952&171.87&0.4026&105051.5 \\ \cline{2-7}
      &color clustering&grouping&0.800&153.76&0.4314&96280.4 \\ \cline{3-7}
      &(our proposal)&grouping+adjusting(location)&0.8095&\textbf{147.54}&\textbf{0.4481}&\textbf{94896.6} \\ \cline{3-7}
      &&grouping+adjusting(location+angle)&\textbf{0.8190}&154.43&0.4255&98808.0 \\ \hline
    \end{tabular}
    }
  \end{center}
\end{table}

Table~\ref{tab2} shows the results of the comparison with images of horizontally placed books.
They were similar to Table~\ref{tab1} except that grouping+adjusting(location+angle) with
R$^2$CNN was worse than the others of our proposal.
This is mainly because some bounding boxes of books next to each other were combined by the nms in the cases
of horizontally placed books.
On the other hand, the detection accuracy of EAST was worse than that of R$^2$CNN, and
we could avoid the nms-caused problem and as a result grouping+adjusting(location+angle) with EAST
could get better performance of BA.

\section{CONCLUSION}\label{concl}
In this paper, we focused on text detection on books which were roughly and disorderly
placed in bookshelves.
We did not train learning-based natural scene text detection methods with appropriate data
but refined the results of them by the grouping and adjusting processes.
Our refinement processes can utilize different learning-based methods without any modification to them.
The experimental results showed that our algorithms work well in most cases under the settings
of the experiments, and we also identified some limitations of the proposed method.

In future work, we will consider detecting texts, which could not be detected by
a learning-based method, by exploiting spine and text strings colors.
We also intend to compare the performance between our proposal and
a learning based model trained with appropriate data of enough amount.

\bibliographystyle{IEEEtran}
\bibliography{ref}

\end{document}